\pdfoutput=1

\documentclass[11pt]{article}

\usepackage{ACL2023}

\usepackage{times}
\usepackage{latexsym}
\usepackage{multirow}
\usepackage{todonotes}
\usepackage{comment}
\usepackage[multiple]{footmisc}
\usepackage{fancyvrb} 
\usepackage[most]{tcolorbox}

\usepackage[T1]{fontenc}

\usepackage[utf8]{inputenc}
\usepackage{todonotes}

\usepackage{microtype}
\usepackage{xcolor}
\usepackage{inconsolata}
\usepackage{algorithm}
\usepackage{algpseudocode}

\usepackage{pifont}
\usepackage[normalem]{ulem}
\usepackage{amsmath}
\usepackage{multirow}
\usepackage{multirow}
\usepackage[normalem]{ulem}

\title{Few-shot Transfer Learning for Knowledge Base Question Answering: \\ Fusing Supervised Models with In-Context Learning }

 \author{Mayur Patidar \\ 
 TCS Research \\
 patidar.mayur@tcs.com 
         \And
         Riya Sawhney \\
         Indian Institute of Technology, Delhi \\
         cs1200374@iitd.ac.in
         \And  
         Avinash Singh \\
         TCS Research \\
         singh.avinash9@tcs.com 
         \And
         Biswajit Chatterjee \\
         TCS Research \\
         chatterjee.biswajit1@tcs.com 
         \AND
         Indrajit Bhattacharya \\ 
         TCS Research \\ 
         b.indrajit@tcs.com 
         \And
         Mausam \\
         Indian Institute of Technology, Delhi \\
         mausam@cse.iitd.ac.in
         }

\author{Mayur Patidar$^\dag$, Riya Sawhney$^\ddag$, Avinash Singh$^\dag$, Biswajit Chatterjee$^\dag$, \\{\bf Mausam$^\ddag$, }{\bf Indrajit Bhattacharya$^\dag$}\\
$^\dag$TCS Research,
$^\ddag$Indian Institute of Technology, Delhi \\
\{patidar.mayur, singh.avinash9, chatterjee.biswajit1, b.indrajit\}
@tcs.com
\,\\ 
riya.sawhney@outlook.com, mausam@cse.iitd.ac.in
}

\begin{document}
\maketitle
\newcommand{\sys}{FuSIC-KBQA}

\begin{abstract}
Existing Knowledge Base Question Answering (KBQA) architectures are hungry for annotated data, which make them costly and time-consuming to deploy. 
We introduce the problem of few-shot transfer learning for KBQA, where the target domain offers only a few labeled examples, but a large labeled training dataset is available in a source domain.
We propose a novel KBQA architecture called \sys{} that performs KB-retrieval using multiple source-trained retrievers, re-ranks using an LLM and uses this as input for LLM few-shot in-context learning to generate logical forms. 
These are further refined using execution-guided feedback.  
Experiments over multiple source-target KBQA pairs of varying complexity show that \sys{} significantly outperforms adaptations of SoTA KBQA models for this setting.  
Additional experiments show that \sys{} also outperforms SoTA KBQA models in the in-domain setting when training data is limited.

\end{abstract}

\section{Introduction}\label{sec:intro}
A Knowledge-base Question Answering (KBQA) system converts a natural language question to a logical query, which is executed on the KB to return the answer. 
The prevalence of diverse KBs, in open domain as well as in enterprises, makes KBQA a task of great practical significance. Unfortunately, traditional supervised models \cite{saxena:acl2022,zhang:acl2022subgraph,mitra:naacl2022cmhopkgqa,wang:naacl2022mhopkgqa,das:icml2022subgraphcbr,ye:acl2022rngkbqa,chen:acl2021retrack,das:emnlp2021cbr,shu-etal-2022-tiara,gu-etal-2023-dont} and even recent LLM-based few-shot in-context learning (FS-ICL) architectures~\cite{li:acl2023kbbinder,nie:arxiv2023kbcoder} require large volumes of labeled (question, logical form) pairs for model training or for exemplar selection. 
Such data is costly and time-consuming to annotate in real applications, and this remains the biggest hurdle for rapid and low-cost deployment of KBQA systems.

A more realistic scenario for a new KB is that a domain expert provides a small number of questions and their logical forms as labels. 
In this paper, we ask whether this small data, in conjunction with a readily available and larger open-domain benchmark KBQA dataset \cite{gu:www2021grailqa,yih:acl2016webqsp,su:emnlp2016-graphqa} seen as a source domain, can be used to bootstrap a KBQA system for the new KB, which forms the target domain. 
We propose this as the \textit{few-shot transfer learning} setting for KBQA. 
To the best of our knowledge, this practically useful setting has not been studied in the KBQA literature. 
It differs from unsupervised transfer learning \cite{cao:acl2022progxfer,ravishankar:emnlp2022}, which assumes a large volume of unlabeled training questions in the target domain.

We first note that source and target tasks can differ along multiple dimensions. First, the two KBs may not have much overlap in schema or data. Additionally, the distributions of logical forms and natural language questions may also be quite different. 
This makes it extremely challenging to transfer source-trained models to a target KBQA system, with limited target supervision.

To handle this challenge, we resort to modern day LLMs, which have achieved unprecedented success in language understanding and code generation tasks, with only few-shot in-context learning (FS-ICL). 
We present \sys{} ({\bf Fu}sed {\bf S}upervised and {\bf I}n-{\bf C}ontext Learning for {\bf KBQA}), which uses source-trained models to construct LLM prompts for FS-ICL. 
In particular, given a target question, multiple source-trained retrievers with complementary strengths retrieve potentially relevant candidate paths, relations and entity types from the KB. 
The LLM then performs re-ranking of this retrieved information to increase their target relevance. 
The re-ranked retrieval is provided in the FS-ICL prompt of another LLM call to generate logical forms. 
To minimize syntax errors, \sys{} generates logical forms in a common language (SPARQL) instead of niche languages used in KBQA (e.g. s-expressions).
To minimize semantic errors, \sys{} provides execution-guided feedback to the LLM based on whether the logical form on execution returns an empty answer.

We perform extensive experiments with multiple pairs of source-target datasets of varying complexity. We show that \sys{} outperforms adaptations of SoTA supervised models and recent few-shot LLM FS-ICL models for KBQA by large margins. Additionally, we find that even for in-domain KBQA (non-transfer) with limited training data, \sys{} outperforms SoTA supervised KBQA models, suggesting that it is robust across various low-data settings.

Our specific contributions are as follows. 
(a) We propose the important task of few-shot transfer learning for KBQA. 
(b) We develop a novel KBQA architecture that combines the benefits of supervised retrievers and LLM prompting for generation and refinement.  
(c) We carefully create multiple KBQA transfer datasets of varying complexity, which we publicly release to aid future research\footnote{\url{https://github.com/dair-iitd/FuSIC-KBQA/}}.
(d) We demonstrate using extensive experiments that our fusion architecture has strong performance for few-shot transfer. 
(e) We additionally demonstrate benefits of this architecture for the in-domain KBQA setting with limited training data.

\section{Related Work}\label{sec:rw}
{\em In-domain supervised KBQA}~\cite{ye:acl2022rngkbqa,chen:acl2021retrack,shu-etal-2022-tiara,gu-etal-2023-dont,faldu:acl2024} involves only a single domain with a large volume of in-domain labeled examples, which are used to train or fine-tune LLM-based architectures.
Almost all SoTA approaches for this problem use a retrieve-then-generate architecture, where the first stage retrieves KB elements relevant to a question, and then provides these as context to a generation stage that uses an LLM such as T5 to generate the logical form.
Pangu~\cite{gu-etal-2023-dont} is different in that it fine-tunes LLMs to discriminate between relevant logical forms, and interleaves retrieval and discrimination step by step.
RetinaQA~\cite{faldu:acl2024} is a contemporaneous KBQA model for answerable as well as unanswerable questions~\cite{patidar-etal-2023-knowledge} that combines traversal-based KB retrieval with sketch-filling based construction. 
It also achieves SoTA performance for in-domain supervised KBQA.

More recent in-domain KBQA approaches use few-shot in-context learning (FS-ICL) with LLMs. 
These are still dependent on large volumes of in-domain training data for selecting the most relevant few-shots.
Different from retrieve-then-generate, KB-Binder~\cite{li:acl2023kbbinder} and KB-Coder~\cite{nie:arxiv2023kbcoder} follow a generate-then-ground architecture that first generates KB-independent partial programs using FS-ICL, and then grounds these for the specific KB using unsupervised techniques.
Crucially, these do not take any advantage of the labeled in-domain examples using supervised learning.
In recent work, \newcite{shu:arxiv2023bottlenecks} use TIARA's supervised retriever followed by FS-ICL LLM generation.
We improve over this basic architecture in multiple ways -- by accommodating multiple retrievers, re-ranking retrieval results and including execution-based feedback.

{\em Unsupervised Transfer Learning for KBQA}~\cite{cao:acl2022progxfer,ravishankar:emnlp2022} involves large volumes of labeled training data in the source domain, and additionally a large volume of {\em unlabeled data} in the target domain.
SoTA models for this setting follow the generate-then-ground architecture, again by first generating KB-independent (partial) programs, which are then adapted for the target KB using unlabeled examples.
These models do not look to benefit from FS-ICL using LLMs.

LLM prompting techniques related to ours have been used outside of KBQA.
Re-ranking of output from supervised or unsupervised retrieval models using LLMs has recently been used for information extraction~\cite{ma2023-large:emnlp-f2023} and multi-hop question answering over documents~\cite{khalifa:acl2023}.
We are not aware of the use of LLMs for retrieval re-ranking for KBQA or for combining rankings from multiple retrievers. 
Reflexion~\cite{shinn:neurips2023reflexion} is a recent RL framework that transforms environment signals to language-based feedback for an LLM. 
It explores python function generation as one of the use-cases, where output accuracy on generated test-cases is fed back to the LLM for improvement.
Our execution-guided feedback can be imagined as a specific instance of this framework, but we are not aware of any use of similar ideas for generating logical forms for KBs.

\section{Problem and Proposed Approach}\label{sec:approach}

We begin by defining the few-shot transfer learning setting for Knowledge Base Question Answering (KBQA), and then present our solution for it.

\paragraph{Few-shot Transfer Learning for KBQA:}
A Knowledge Base $G$ consists of a schema with data stored under it.
The schema consists of entity types $T$ and binary relations $R$ defined over pairs of types. 
The data consists of entities $E$ as instances of types $T$, and triples or facts $F\subseteq E \times R \times E$.
We are given a {\em target Knowledge Base} $G^t$ (consisting of entity types $T^t$, relations $R^t$, entities $E^t$ and facts $F^t$) and a natural language question $q^t$, and the goal is to generate a structured query or a logical form $l^t$, which when executed over $G^t$ returns a non-empty answer $A^t$ for the question $q^t$.

In few-shot transfer learning, we are provided with {\em target few-shots} $D^t$ containing tens of labeled training examples of questions and logical forms in the target domain.
In addition, a related source domain has a {\em source knowledge base} $G^s$ consisting of its own types $T^s$, relations $R^s$, entities $E^s$ and facts $F^s$, and a much larger {\em source training set} $D^s$ containing thousands of labeled training examples of questions and corresponding logical forms.
A special case of this setting is {\em zero-shot transfer learning}, where  
$D^t=\{\}$.

\begin{figure*}[t]
    \centering
    \includegraphics[width=0.86\textwidth]{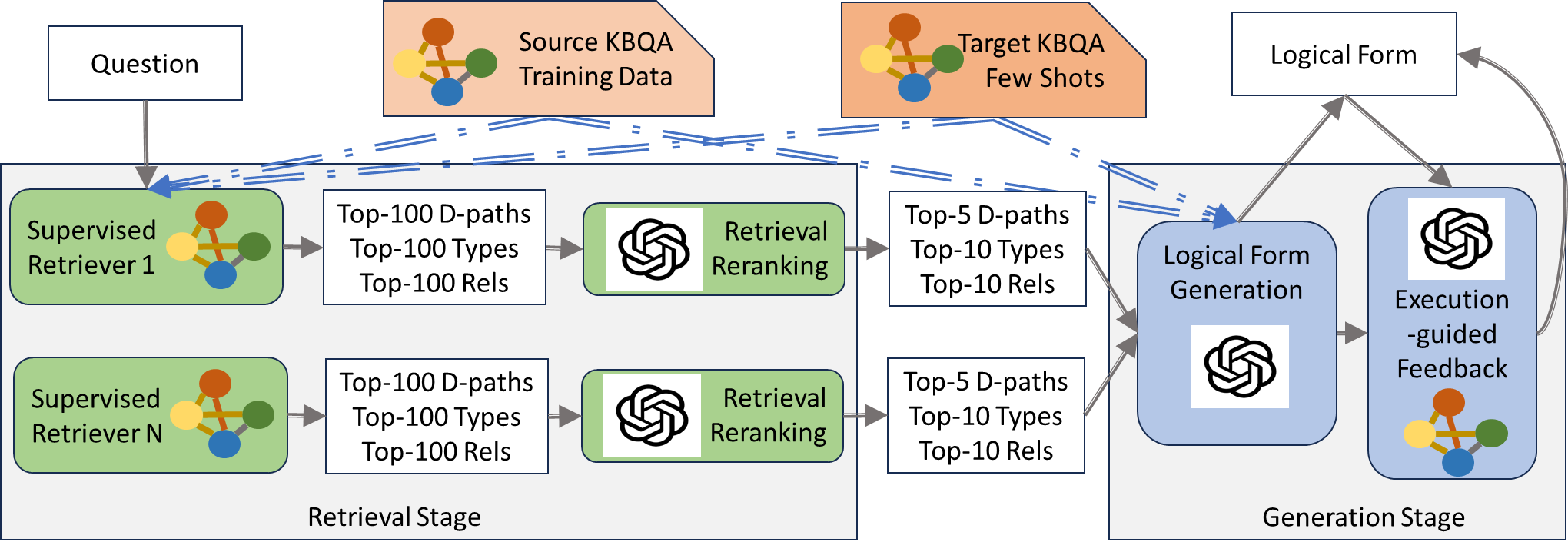}
    \caption{Architecture of \sys{}, which fuses supervised learning with LLM few-shot in-context learning for KBQA. Starting from the question, the Retrieval Stage retrieves KB elements using one or more supervised KB retrievers and re-ranks these using LLM prompting. The subsequent Generation Stage uses LLM prompting with the retrieved KB context and labeled exemplars to generate the logical form, which is refined again with LLM prompting, using feedback from execution over the KB. The question is an input to all the components. Blue arrows with compound lines show flow of source training data and target few shots to components.}
    \label{fig:arch}
\end{figure*}

\paragraph{Proposed Approach:}
{\em Retrieve-then-generate} architectures for KBQA -- used by state-of-the-art models for the supervised in-domain setting -- first retrieve relevant KB-elements for the query and then use these as context for generating the logical form. 
The KB-retrievers used in such models have steadily grown in sophistication in recent years.
On the other hand, {\em generate-then-ground} models for KBQA -- used for unsupervised transfer learning and in domain few-shot learning with LLMs -- first generate KB-independent incomplete or draft logical forms based on the question, and then ground these for the specific KB.

For our solution, we assume that we have access to a Black-box Large Language Model. 
We use this term to refer to 100B+ parameter LLMs such as GPT-3 and beyond, which cannot be trained or fine-tuned for all practical purposes, and only offer access through prompting and in-context learning.
In the rest of this paper, when we refer to LLMs, unless otherwise specified, we mean such models.

We begin by asking: 
{\em which of the two KBQA architectures above is better suited to simultaneously take advantage of a large source training set and an LLM to accurately generate logical forms with limited target supervision?}
LLMs excel at understanding and generating language, but lack knowledge of the target KB.
Therefore, we hypothesize that an LLM can generate complete logical forms accurately using in-context exemplars from the target, when provided with high quality context retrieved from the target KB. 
High quality retrieval can be achieved using one or more sophisticated supervised retrievers -- trained using the large source training set, and optionally adapted for the target using few-shot fine-tuning -- and then re-ranking for the target using LLM prompting.
Based on this hypothesis, we propose \sys{}, a retrieve-then-generate architecture for the few-shot transfer learning problem for KBQA. 
It is depicted at a high-level in Fig.~\ref{fig:arch}.

\paragraph{Retrieval Stage:} 
The retrieval stage makes use of one or more off-the-shelf KB-retrievers.
A KB-retriever takes as input a question $q$ and a KB $G$ and returns 3 different aspects: (i) top-$K$ data-paths starting from the linked entities in $q$, (ii) top-$K$ entity types, and (iii) top-$K$ relations.
Not all retrievers may return all three aspects, and $K$ may be aspect-specific for a retriever.
We additionally assume that the retrievers are supervised and allow training and fine-tuning.
The retrievers of SoTA KBQA models such as TIARA~\cite{shu-etal-2022-tiara}, RnG-KBQA~\cite{ye:acl2022rngkbqa} and ReTRaCK~\cite{chen:acl2021retrack} satisfy these requirements. Though Pangu~\cite{gu-etal-2023-dont} and RetinaQA~\cite{faldu:acl2024} do not have explicit retrievers, these can be adapted as such.
More details on TIARA and Pangu retrievers are in Sec. ~\ref{subsubsection:retriever}.

We train each retriever on source data $D^s$ and fine-tune using target few-shots $D^t$. 
For a target test question, \sys{} first performs retrieval using each retriever to get top-$K$ results for each aspect. 
Note that $K$ may be aspect-specific.
Next, the target relevance is improved by {\em retrieval re-ranking} and sub-selecting using LLM prompting. 
Specifically, top-$K$ results for each aspect are provided to the LLM along with the question, and it is prompted to select and rank the top-$k$ results ($k\ll K$).
Note that $k$ may also be aspect-specific.
Aspect-wise top-$k$ retrieval from each retriever is separately passed to the generation stage.
Specific prompts are described in Sec. \ref{subsec:prompts_for_retrieval_stage}.

\paragraph{Generation Stage:}
We use LLM prompting to generate target logical forms. 
The prompt includes (i) the test question, (ii) the retrieved KB context for each aspect of each retriever, (iii) an instruction to generate the logical form, and (iv) few-shot exemplars from the target.
For zero-shot, target exemplars are replaced with randomly selected exemplars from the source training set $D^s$. 
Example prompts are included in Sec.~\ref{subsec:prompts_for_generation_stage}.
Different retrievers are not only allowed to have different retrieval sizes ($k$) for the same aspect, but also different output structures.
For example, one retriever may return relations while another may not.
Including complete retrieval outputs from multiple retrievers in the generation prompt may not be possible due to length constraints when using few-shots, since the retrieval is included for the exemplars as well.
While optimal merging is a challenging problem, in this work, we linearly decrease the retrieval size $k$ for aspects of individual retrievers, as the number of retrievers increases.
Sec.~\ref{subsec:prompts_for_generation_stage} has more details.

Even with few-shot exemplars and high quality context from multiple retrievers, LLM-generated logical forms contain syntactic and semantic errors, because of unfamiliarity with the target questions and logical forms.
Choice of the representation language significantly impacts syntactic correctness of the generated logical form. 
So far, s-expressions have been used for LLM-based generation~\cite{li:acl2023kbbinder,nie:arxiv2023kbcoder,shu:arxiv2023bottlenecks}. 
We expect any LLM to be significantly more familiar with SPARQL than niche languages such as s-expression, and therefore be better able to generate syntactically correct programs. 
We therefore use SPARQL.
SoTA KBQA models represent retrieved KB data paths also as programs. 
For representational consistency, 
we use SPARQL to represent the retrieved KB data paths as well.

Aside from syntactic errors, the generated logical form may have KB-specific semantic errors. Supervised KBQA models commonly use execution-guided checks to filter the ranked list of generated logical forms~\cite{shu-etal-2022-tiara,gu-etal-2023-dont,faldu:acl2024} by checking if a logical form upon execution returns an empty answer.
Instead, we use LLM-based {\em execution-guided refinement}.
Specifically, if an empty answer is obtained on executing the generated logical form, this feedback is verbally provided to the LLM along with the erroneous logical form, prompting it to correct itself.
This is continued until a non-empty answer is obtained, or for a maximum number of iterations, each time providing the sequence of previous generations as context.
Examples of the specific prompt are in Sec. \ref{subsec:prompts_for_execution_based_feedback}.
\section{Experiments} \label{sec:experiments}

We now present experimental evaluation of \sys{} for few-shot transfer learning which is our main focus.
We also evaluate its usefulness for the limited training data in-domain KBQA setting.

\subsection{Experimental Set-up}\label{subsec:exp-setup}

\begin{table*}[ht]
\begin{center}
\small
            \begin{tabular}[b]{lccccccc}
\hline
 \multirow{3}{3em}{\textbf{Source$\rightarrow$Target}}   & \multicolumn{2}{c}{\textbf{Kn. Base}} & \multicolumn{2}{c}{\textbf{Logical Forms}} &\multicolumn{2}{c}{\textbf{NL Questions}}& \textbf{Target}  \\
  & Dom. & Rel. & Func. & \# Relations &Src-Tgt Sim &\# Tokens & \textbf{Gen.} \\  
  & JS & \% New & JS & Src Avg/Tgt Avg &Cosine &Src Avg/Tgt Avg & \textbf{Scenario} \\  
\hline
GrailQA$\rightarrow$GraphQA &0.25 &50.4 &0.01 & 1.36 / 1.53 &0.44 & 10.4 / 9.3 & C,Z\\ 
GrailQA$\rightarrow$WebQSP &0.37 &55.4 &0.04 & 1.35 / 1.54 &0.31 & 10.4 / 6.6 & I,C,Z\\ 
WebQSP$\rightarrow$GrailQA-Tech &0.68 &99.3 &0.11 & 1.52 / 1.46 &0.23 & 
6.6 / 10.9 & I,C,Z\\ 
WebQSP$\rightarrow$GraphQA-Pop &0.66 &97.5 &0.07 & 1.52 / 1.73 &0.23 & 6.6 / 9.2 & C,Z\\ 
\hline
\end{tabular}
\end{center}
\caption{Statistics for different source and target (test set) KBQA task pairs in terms of the knowledge base, logical forms and natural language questions. `Dom.(JS)' is Jensen Shannon (JS) divergence between domain distribution of questions, `Rel.(\% New)' is percentage of questions in target with new (unseen) relations, `Func.(JS)' is JS-divergence between distributions over functions in logical forms, `\# Relations' shows the source average and the target average for number of relations per logical form, `Src-Tgt Sim' is average minimum cosine distance between source and target questions and `\# Tokens' shows the source average and target average of number of tokens per question. For the generalization setting of the target test set wrt target few shots, I,C and Z denote IID, compositional generalization and zero-shot respectively.} 
\label{tab:datasets}
\end{table*}

\paragraph{Datasets:}
A crucial aspect is creating source and target KBQA task pairs that are challenging for few-shot transfer in different ways.
Source and target KBQA tasks can differ along multiple dimensions: (i) KB schema and data, (ii) distribution of logical forms, and (iii) distribution of natural language questions. 
An additional source-independent dimension is (iv) the generalization scenario for the target test set versus the target few shots.  
We prepare source$\rightarrow$target pairs using the benchmark KBQA datasets GrailQA~\cite{gu:www2021grailqa}, WebQSP~\cite{yih:acl2016webqsp} and GraphQA~\cite{su:emnlp2016-graphqa}.
Though all of these have Freebase as the backbone KB, there is large divergence between the domains, relations and types that these span.
GrailQA and GraphQA were constructed similarly by first sampling meaningful KB paths and then verbalizing these into questions.
However, the GrailQA test set contains largely IID (seen schema elements) and some compositional (unseen combination of seen schema elements) and zero-shot questions (unseen schema elements), while GraphQA test questions are all compositional and therefore harder.
WebQSP questions are popular web queries posed by real users, and are therefore significantly different from GrailQA and GraphQA questions.
However, all its test questions are IID.

Using these benchmark datasets, we created four source$\rightarrow$target pairs in two ways. 
First, using entire benchmark datasets as sources and targets, we created GrailQA$\rightarrow$GraphQA and GrailQA$\rightarrow$WebQSP, considering GrailQA's large training dataset size (>44K).
Then we took WebQSP as the source and selected as two targets \textit{subsets} of GrailQA domains (in the technical category) and GraphQA domains (in the popular category) that are minimally represented in the WebQSP training data.
We name these targets GrailQA-Tech and GraphQA-Pop.
Since many functions are not present in WebQSP logical forms, WebQSP$\rightarrow$GrailQA-Tech and WebQSP$\rightarrow$GrailQA-Pop present significant transfer challenges.
Tab.~\ref{tab:datasets} records some key statistics of the four source$\rightarrow$target pairs.
We can see that while all datasets pose challenges for few-shot transfer, overall difficulty increases from top to bottom.
Because of cost of experimentation with GPT-4, for each target, we create the target test sets by sampling 500 instances uniformly at random from the original test sets (dev set for GrailQA) restricted to selected domains as necessary.
100 target few-shots are similarly sampled from the corresponding training sets. 
More details about dataset construction are in Sec.~\ref{subsec:dataset_construction_details}.

\paragraph{LLM and Supervised Retrievers:}
We use \texttt{gpt-4-0613} as the LLM, since recent experiments have demonstrated GPT-4's significant superiority across a wide range of NLP tasks~\cite{bubeck2023sparks}.
As supervised retrievers, we experiment with TIARA and Pangu.
TIARA~\cite{shu-etal-2022-tiara} has the most general and sophisticated retriever among SoTA supervised KBQA models.
Its retriever generalizes that of RnG-KBQA~\cite{ye:acl2022rngkbqa} and ReTRaCK~\cite{chen:acl2021retrack} and returns data paths, types and relations.
Though Pangu~\cite{gu-etal-2023-dont} does not have an explicit retriever and generates its logical form by incrementally selecting relations, we adapt it as a retriever by taking its top-5 paths at the final stage and extracting relations and types from these.
We use TIARA's entity linker (trained on GrailQA) as our pre-trained off-the-shelf entity-linker for all experiments.
We equip all retrieve-then-generate baselines with this same entity linker.
KB-Binder's generate-then-ground architecture uses its own LLM-based entity linker.

\paragraph{Baselines:}
Since there are no existing few-shot transfer KBQA models, we adapt SoTA KBQA models designed for the in-domain setting from the purely supervised and LLM FS-ICL categories. 

\noindent \emph{Supervised KBQA models:} Considering the GrailQA leader-board and availability of code, we use \textbf{Pangu}~\cite{gu-etal-2023-dont} and \textbf{TIARA}~\cite{shu-etal-2022-tiara} as supervised KBQA models. 
We use publicly available code, and pre-trained check-points for our sources - GrailQA and WebQSP. 
We additionally experiment with \textbf{BERT-Ranking}, which has been shown to work well in a different transfer setting for KBQA (non few-shot, with large volumes of target training data)~\cite{gu:www2021grailqa,shu-etal-2022-tiara} and outperform TIARA~\cite{shu:arxiv2023bottlenecks}.
We fine-tune all trainable components of these models on target few-shots for 1 epoch.
These models are customized for generating logical forms in s-expression, and we preserve this in our experiments.  Sec.~\ref{subsec:supervised_model_details} has more details.

\noindent \emph{LLM Few-shot ICL KBQA models:} 
We use \textbf{KB-Binder}~\cite{li:acl2023kbbinder} as the SoTA model in this category.
In the retrieval (-R) setting, KB-Binder uses BM25 to retrieve the demonstration samples from the entire available training data.
We restrict its retrieval to the target few-shots in the few-shot setting and source training data in the zero-shot setting.
We use self-consistency and majority voting with 6 examples, as in the paper. 
Additionally, we experiment with a recent retrieve-then-generate LLM FS-ICL model~\cite{shu:arxiv2023bottlenecks} which we name as \textbf{gf-LLM} (ground-first LLM).
As in the paper, we keep TIARA as its retriever, but pre-trained on the source and fine-tuned on target few-shots as in \sys{}.
The in-context examples are provided in the same way as KB-Binder.
KB-Binder and gf-LLM report experiments using \texttt{code-davinci-002} and \texttt{gpt-3.5-turbo-0613} respectively.
We replace these with \texttt{gpt-4-0613} as in \sys{}.
Like Pangu and TIARA, these models also generate logical forms in s-expression, which we preserve.
Note that unlike KB-Binder, \sys{} and gf-LLM include KB-retrieval in their generation prompts for the test question and the exemplars, requiring many more tokens per exemplar.
For all three models, we provide the maximum number of exemplars that fit in the prompt -- 100 for KB-Binder, and 5 for \sys{} and gf-LLM.
Sec.~\ref{subsec:llm_baseline_details} has more details.

\begin{table*}[ht]
\begin{center}
\small
            \begin{tabular}[b]{lllcccc}
\hline
& & \multirow{2}{4em}{\textbf{Model}}   & \textbf{GrailQA$\rightarrow$} &\textbf{GrailQA$\rightarrow$} &\textbf{WebQSP$\rightarrow$} &\textbf{WebQSP$\rightarrow$}\\  
& &  & \textbf{GraphQA} &\textbf{WebQSP} &\textbf{GrailQA-Tech} &\textbf{GraphQA-Pop}\\  
\hline
\multirow{8}{1em}{\rotatebox[origin=c]{90}{Zero-shot}} &  \multirow{3}{5em}{Supervised} & TIARA~\cite{shu-etal-2022-tiara} &52.8 &30.0 &30.6 & 37.8\\
  & & Pangu~\cite{gu-etal-2023-dont} &60.9 &39.3 & 40.8 &25.1\\ 
  & & BERT-Ranking~\cite{gu:www2021grailqa} &42.4 &35.9 &21.5 & 21.1\\ 
& \multirow{1}{5em}{LLM ICL} & KB-Binder~\cite{li:acl2023kbbinder} &45.0 &39.0 & 33.6 &25.1 \\
& \multirow{4}{5em}{Fusion} & gf-LLM~\cite{shu:arxiv2023bottlenecks} &53.4 &31.3 &42.8 &38.1 \\
  &  & \sys{}(T) &59.2 &55.9 &68.4 & 56.8 \\
  &  & \sys{}(P) & 63.8 &52.5  &65.1 &41.3 \\
  &  & \sys{}(T,P) & \textbf{67.5} &\textbf{56.6}  & \textbf{71.4}& \textbf{61.7}\\
\hline
\multirow{8}{1em}{\rotatebox[origin=c]{90}{Few-shot}} &  \multirow{3}{5em}{Supervised} & TIARA~\cite{shu-etal-2022-tiara} &50.1 &40.6  & 48.2&37.3 \\
  & & Pangu~\cite{gu-etal-2023-dont} &63.2   &45.3 & 51.0 & 36.7\\ 
  & & BERT-Ranking~\cite{gu:www2021grailqa} &44.1    &34.4 & 33.1&23.4 \\ 
& \multirow{1}{5em}{LLM ICL} & KB-Binder~\cite{li:acl2023kbbinder} &35.2 &55.7 & 68.0 &31.5\\
    & \multirow{4}{5em}{Fusion} & gf-LLM~\cite{shu:arxiv2023bottlenecks} &52.5 &53.4 & 61.2&46.6\\
  &  & \sys{}(T) &60.3 &65.0 &70.8  &52.3\\
   &  & \sys{}(P) & 64.5 & 61.9 &70.3 & 49.1 \\
    &  & \sys{}(T,P) & \textbf{69.1} & \textbf{65.1} &\textbf{74.6} &\textbf{61.6}\\
\hline
\end{tabular}
\end{center}
\caption{Few-shot transfer learning performance (F1) of different models for different source$\rightarrow$target pairs, evaluated over 500 randomly sampled instances over the test sets. \sys{}(T), \sys{}(P), \sys{}(T,P) denote \sys{} used with TIARA, Pangu, and TIARA and Pangu together as supervised retrievers.  
}
\label{tab:transfer}
\end{table*}

\begin{table}[ht]
\begin{center}
\small
            \begin{tabular}[b]{lllllc}
\hline
& \multirow{2}{1em}{\textbf{Model}}   & \textbf{Gl$\rightarrow$} &\textbf{Gl$\rightarrow$} &\textbf{W$\rightarrow$}&\textbf{W$\rightarrow$}  \\  
&  & \textbf{Gp} & \textbf{W} &\textbf{Gl-T}&\textbf{Gp-P}\\
\hline
\multirow{4}{1em}{\rotatebox[origin=c]{90}{Zero-shot}}  &  \sys{}(T) &59.2 &55.9 &68.4&56.8\\ 
 &  w/o SPARQL &48.2 & 41.2 &60.5& 46.4\\
  &  w/o EGF &55.1 &53.2&64.4&54.9\\ 
&  w/o RR & 58.0 & 48.0 & 59.5&52.6\\ 
\hline
\multirow{4}{1em}{\rotatebox[origin=c]{90}{Few-shot}}    &  \sys{}(T) &60.3  &65.0&70.8&52.3\\ 
  &  w/o SPARQL &53.0 & 58.7 &64.3&53.9\\ 
  &  w/o EGF &56.5 & 62.0 &66.6&44.0\\ 
&  w/o RR &53.7 &  61.2 &68.4&50.5\\
\hline
\end{tabular}
\end{center}
\caption{Ablation of few-shot and zero-shot transfer learning performance using F1 for \sys{}(T) for different source$\rightarrow$ pairs. 
Gl, Gl-T, Gp, Gp-P and W denote GrailQA, GrailQA-Tech, GraphQA, GraphQA-Pop and WebQSP, respectively, for brevity.
EGF and RR denote execution-guided feedback and retrieval reranking respectively. w/o SPARQL denotes use of s-expression instead of SPARQL for generation.
}
\label{tab:transfer-ablation}
\end{table}

\subsection{Results for Few-shot Transfer Setting}

For the few-shot transfer learning, we first compare \sys{} with SOTA baselines and then analyze different aspects of the model. 

\subsubsection{Model Comparison} Performances of various models for the few-shot transfer setting are recorded in Tab.~\ref{tab:transfer}.
Because of mismatch between logical form language of some models and the gold-standards, we directly evaluate the answers obtained upon logical form execution using F1~\cite{gu-etal-2023-dont,li:acl2023kbbinder}. 

We first make some key observations about the performance of \sys{}.
{\bf (a)} \sys{} performs the best by a large margin in all datasets for both few-shot and zero-shot.
{\bf (b)} Performance varies with the retriever -- \sys{} with TIARA is mostly better than \sys{} with Pangu, but using the two together is always better than both individually, showing the complementary nature of their retrieval and \sys{}'s ability to take advantage of this.
{\bf (c)} While performance improves with few-shots, it also performs remarkably well in the zero-shot setting across datasets. 
{\bf (d)} Performance varies for all models across datasets, as a reflection of their varying difficulty, and the fourth dataset is the hardest for almost all models.
However, \sys{} has the most stable performance across datasets.

Next, we analyze relative performance of the different models.
{\bf (a)} Performances of the supervised models degrade significantly as task complexity increases. While these, especially Pangu, perform very well for the first dataset, performance is significantly lower for the other three. 
This reflects the challenge of learning to generate logical forms accurately from only target few-shots when the source is less related.
{\bf (b)} Models equipped with LLM FS-ICL perform better than purely supervised models, particularly with few-shots. 
This demonstrates the generative ability of GPT-4 ICL for logical forms. 
That said, the naive LLM FS-ICL baseline, which simply prompts GPT-4 with a question without any KB retrieval and instructs it to generate the logical form along with 5 question and logical form exemplars, performs extremely poorly, e.g. achieving less than 5.0 F1 on WebQSP$\rightarrow$GrailQATech.
{\bf (c)} The retrieve-then-generate approach of gf-LLM and \sys{} works better than that of KB-Binder's generate-then-ground approach, demonstrating the strength of this architecture for few-shot transfer.
{\bf (d)} Within the retrieve-then-generate category, \sys{} significantly outperforms gf-LLM, by accommodating multiple retrievers, using SPARQL and employing LLMs for retrieval re-ranking and execution guided refinement in addition to generation.

\subsubsection{Analysis of \sys{}}
\paragraph{Ablations:} Tab.~\ref{tab:transfer-ablation} records performance for different ablations of \sys{}.
We can see that the biggest gain comes from using SPARQL as the logical form language, except for the few-shot setting in the fourth dataset.
Refinement using execution-guided feedback and reranking of retrieval results make more modest but significant contributions. 

We performed additional ablation experiments on the  WebQSP$\rightarrow$GrailQATech dataset with \sys{(T,P)}, (i.e., with both TIARA and Pangu as retrievers).
We first varied the number of EGF iterations.  
In the few-shot setting, performance without EGF is 71.6 F1. This improves to 74.4 and 74.6 after the first and second EGF iterations, but does not improve thereafter. In the zero-shot setting, in contrast, performance steadily improves from 66.7 F1 before EGF to 68.6, 70.8, 71.2 and 71.4 over four EGF iterations.
This demonstrates that while each EGF iteration has associated cost, these may bring some performance benefits.

For the same WebQSP$\rightarrow$GrailQATech dataset again using \sys{}(T,P), we studied the effect of the number of few shots.
Recall that each shot contributes significantly to prompt length.
Unexpectedly, we observed F1 to be 73.6, 70.5, 72.3 and 74.6 for zero, 1, 3 and 5 shots respectively.
The poorer performance of 1 and 3 shots can be attributed to low exemplar diversity, biasing GPT-4 to predict simpler logical forms.  
These suggests that \sys{} could be used with fewer shots, but diversity of exemplars is important.

\paragraph{Error Analysis:} We performed detailed error analysis of the best performing version of \sys{}, that with TIARA and Pangu as retrievers, for the GrailQA$\rightarrow$WebQSP and WebQSP$\rightarrow$GrailQA-Tech datasets in the few-shot setting.
For GrailQA$\rightarrow$WebQSP, 41.6\% of test questions have errors in answers (i.e., F1 $<$ 1.0). 
Of these, 17.4\% are due to entity linking, 9.8\% due to retrieval and 14.4\% due to generation.
On the other hand, for WebQSP$\rightarrow$GrailQA-Tech, there are fewer errors in total (31\%), and the corresponding percentages are 10\%, 0.4\% and 21\%, implying that entity linking errors are less, while all other errors are essentially generation errors -- there are almost no retrieval errors. 
Entity linking errors are caused roughly in equal proportion by errors in mention detection and entity disambiguation -- 8.6\% and 8.8\% in the first dataset, and 4\% and 6\% in the second. 
Note that all models other than KB-Binder use the same entity linker.
Retrieval errors can be of 3 types -- (i) correct KB elements not in the top-$K$ of the original retrieval, (ii) in the top-$K$ but not in the top-$k$ (with $k\ll K$) of the original retrieval, and (iii) originally in the top-$k$, but incorrectly demoted by LLM re-ranking. 
For the first dataset, which has significant retrieval errors, these contribute to 3.6\%, 5.8\% and 0.4\% of errors -- indicating that re-ranking errors are negligible. 
Finally, generation errors are potentially of two kinds -- syntactic and semantic. 
However, for both datasets, all generation errors (14.4\% and 21\%) are semantic, with no syntactic errors at all.

\begin{table}[ht]
\begin{center}
\small
            \begin{tabular}[b]{lllllc}
\hline
& \multirow{2}{1em}{\textbf{LLM}}   & \textbf{Gl$\rightarrow$} &\textbf{Gl$\rightarrow$} &\textbf{W$\rightarrow$}&\textbf{W$\rightarrow$}  \\  
&  & \textbf{Gp} & \textbf{W} &\textbf{Gl-T}&\textbf{Gp-P}\\
\hline
\multirow{2}{1em}{\rotatebox[origin=c]{90}{Z-s}}  &  GPT-4 &67.5 &56.6 &71.4&61.7\\ 
 &  Mixtral &48.9 & 36.8 &59.9& 41.4\\
\hline
\multirow{2}{1em}{\rotatebox[origin=c]{90}{F-s}}    &  GPT-4 &69.1  &65.1&74.6&61.6\\ 
  &  Mixtral &54.0 & 48.2 &59.0&34.2\\ 
\hline
\end{tabular}
\end{center}
\caption{Performance of \sys{}(T,P) with GPT-4 and Mixtral for the four datasets named as in Tab.~\ref{tab:transfer-ablation}.}
\label{tab:open-source-llm}
\end{table}

\paragraph{Generalization for \sys{}: }
In experiments so far, we have used GPT-4 as the state-of-the-art LLM for \sys{}. 
Further, the source and target in all four datasets, in spite of their significant differences, have the same KB (Freebase) as the backbone. 
We now report some experiments with a different LLM and a different KB.

A major strength of \sys{} is that it is a framework that allows plugging in different retrievers and also different LLMs. 
While we have demonstrated its superior performance using GPT-4, GPT-4 is expensive and proprietary.
So we experimented with Mixtral 8x7B Instruct v0.1\footnote{\url{https://huggingface.co/mistralai/Mixtral-8x7B-Instruct-v0.1}}, which is a state-of-the-art open-source LLM with a large context length (32k), which is important for \sys{}.
Results are recorded in Tab.~\ref{tab:open-source-llm}.
Performance of FuSIC-KBQA(T,P) with Mixtral drops significantly compared to GPT-4, as expected. 
However, comparing with Tab.~2, we see that \sys{} with Mixtral outperforms the supervised baselines for two out of the four datasets in both the few-shot and the zero-shot settings.
This demonstrates that \sys{} holds promise even with significantly smaller open-source LLMs. More details are in Sec. \ref{subsec:open-source-llm}

\begin{table}[ht]
\begin{center}
\small
            \begin{tabular}[b]{lccc}
\hline
& \textbf{TIARA}   & \textbf{gf-LLM} &\textbf{\sys{}}  \\  
\hline
{\bf GrailQA$\rightarrow$} & \multirow{2}{1em}{5.7} &\multirow{2}{1em}{13.8}&\multirow{2}{1em}{\textbf{45.6}}\\
{\bf MAKG} &  & & \\
\hline
\end{tabular}
\end{center}
\caption{Performance (F1) of \sys{}, gf-LLM and TIARA in the few-shot setting with GrailQA as  source and Microsoft Academic Graph (MAKG) as target. Since the complete KB schema fits in the prompt, \sys{} does not use a retriever.}
\label{tab:makg}
\end{table}

We also created a small new target dataset using the Microsoft Academic KG\footnote{\url{(https://makg.org/)}} 
This has real questions from students, which we manually annotated with corresponding SPARQL queries. 
We divided 70 questions into 5 training examples and 65 test questions. 
We used GrailQA as the source.
The results are recorded in Tab.~\ref{tab:makg}.
We can see that \sys{} outperforms TIARA and also gf-LLM by huge margins in this challenging setting.
While this is a small experiment, this demonstrates the \sys{} is likely to generalize well for different source and target backbone KBs.
More details are in Sec.~\ref{subsec:makg_details}.

\begin{table}[h]
\begin{center}
\small
            \begin{tabular}[b]{clcc}
\hline

  \multicolumn{1}{c}{\textbf{\#Train samples}}&\multicolumn{1}{l}{\textbf{Model}}&\textbf{F1} & \textbf{EM} \\ \hline
  \multirow{5}{*}{ 2,217 (5\%)}&TIARA & 64.8 &58.2  \\ 
  & Pangu  & 69.5& 62.8\\
  & RetinaQA &62.4 &57.0 \\
  & \sys{}(T)  & 70.6 & 60.4 \\
  & \sys{}(P)  & 72.9 &62.0 \\
   & \sys{}(T,P)  & \textbf{74.5} & \textbf{63.8}\\
  \hline
   \multirow{5}{*}{4,434 (10\%)}&TIARA &67.2& 61.2  \\ 
  & Pangu  & 72.4& 64.6\\
  & RetinaQA &68.3& 62.4\\
   & \sys{}(T) & 74.8 & 63.0\\
    & \sys{}(P) & 74.1 & 63.0\\
    & \sys{}(T,P)  & \textbf{75.6} &\textbf{65.2} \\
  \hline
  \multirow{5}{*}{ 44,337 (100\%)}&TIARA &82.6& 76.0  \\ 
  & Pangu  & 83.5& 76.2\\
  & RetinaQA & \textbf{84.7}& 78.4\\
  & \sys{}(T) & 79.1 &70.2 \\
  & \sys{}(P) & 82.5 &72.8 \\
  & \sys{}(T,P)  & 83.6 & 75.0\\
  \hline
 \end{tabular}
\end{center}
\caption{In-domain performance of different models on randomly sampled 500 questions from GrailQA dev set, when trained using different percentages of the GrailQA training set. TIARA, Pangu and RetinaQA generate s-expressions while \sys{} generates SPARQL, so that EM is measured differently.}
\label{tab:in-domain}
\end{table}

\subsection{Results for In-domain Setting}

Finally, we demonstrate the usefulness of \sys{} in the in-domain setting when the volume of training data is limited.
In addition to TIARA and Pangu, we also compare against RetinaQA~\cite{faldu:acl2024} -- a contemporaneous model that has achieved SoTA performance for this setting. 
We train the supervised KBQA models and also the supervised components of \sys{} on progressively smaller subsets of the GrailQA training data, and evaluate their performance on 500 randomly chosen instances from the GrailQA dev set.

Here, we additionally evaluate logical form accuracy using exact match (EM).
Since the supervised models generate s-expressions, we evaluate EM automatically~\cite{ye:acl2022rngkbqa}.
\sys{} generates SPARQL. 
Since automatic equivalence checking of SPARQL programs is challenging, we adopt a partly automated approach, where the checker outputs \textsc{equivalent}, \textsc{non-equivalent} or \textsc{no-decision}, and candidates in the last category are evaluated manually.
A sample-based assessment on the test set showed a precision of 1.0 for the first two decisions.
Further details are in Sec.~\ref{subsec:em-sparql}.

The results are recorded in Tab.~\ref{tab:in-domain}. We can see that with 5\% and 10\% training data, \sys{}(T) outperforms TIARA in terms of F1 and EM, \sys{}(P) outperforms Pangu in terms of F1 but not EM, but \sys{}(T,P) outperforms all the supervised models in terms of EM and also by a large margin in terms of F1.
Even with 100\% training data, \sys{}(T,P) outperforms TIARA and Pangu in terms of F1.
However, in this setting, RetinaQA is the best model overall in terms of both F1 and EM.
It should be noted that in principle \sys{} can be augmented with RetinaQA also as one of the retrievers just as with Pangu, and this could boost its performance.
We leave this for future experiments.

In summary, this experiment demonstrates that \sys{} can outperform supervised models in the in-domain setting given limited training data, and compares favorably even when large volumes of training data are available.

\section{Discussion}
\label{sec:discussion}

Our exploration of LLM for KBQA led to many failed experiments.
We discuss these briefly here. 

\textit{Mention detection and entity disambiguation:} 
We experimented with FS-ICL with GPT-4 for mention detection.
However, this performed worse than the supervised version both in the transfer setting (by 1 pct point) and in the limited training data (5\%) in-domain setting (by 10 pct points).
For entity disambiguation, we explored GPT-4 re-ranking of the entities returned by the supervised model.
However, this did not improve performance.

\textit{Self-consistency and majority voting:} 
This is reported to improve performance for in-domain KBQA with chatGPT~\cite{li:acl2023kbbinder,nie:arxiv2023kbcoder}.
However, with GPT-4, generating with higher temperature led to little variance in the generated logical forms in our experiments, naturally rendering majority voting meaningless.

\textit{Exemplar selection:} 
KB-Binder~\cite{li:acl2023kbbinder} and gf-LLM~\cite{shu:arxiv2023bottlenecks} benefit from dynamic exemplar selection using BM25. 
However, we observed only minor improvements beyond random exemplar selection and static exemplars. 
\section{Conclusions}
\label{sec:conclusion}

We have motivated and addressed the few-shot transfer learning setting for KBQA.
We propose \sys{} as a retrieve-then-generate architecture that benefits from KB-retrieval using one or more source-trained supervised retrievers, which are adapted for the target using LLM-based re-ranking. 
Such retrieval provides precise and relevant context to an LLM that generates logical forms with few-shot in-context learning, further refined by execution-guided feedback. 
We demonstrate the usefulness of \sys{} using carefully designed experiments over multiple source$\rightarrow$target settings of varying complexity. 
We show that \sys{} significantly outperforms adaptations of SoTA KBQA models, both supervised and using LLM few-shot in-context learning, for few-shot transfer.
We also demonstrate that \sys{} outperforms SoTA supervised KBQA models in their own backyard -- the in-domain setting -- when the volume of training data is limited.
An interesting future direction is addressing unanswerability of questions in the few-shot transfer setting~\cite{patidar-etal-2023-knowledge,faldu:acl2024}.
\section*{Limitations}\label{sec:limitations} 

\sys{} makes more LLM API calls compared to the LLM-based baselines. 
Both KB-Binder and gf-LLM make only one API call per test instance.
\sys{} makes 3 additional API calls for re-ranking the 3 different aspects of the retrieval result. 
Additionally, the generator makes 0 to 4 additional API calls per test instance for execution-guided feedback. 
As an example, the average number of EGF-related API calls per instance is 0.7 for GrailQA$\rightarrow$WebQSP.
So, in aggregate, \sys{} makes 4.7 API calls on average for each instance.
The number of API calls increases with additional retrievers.
The goal of our exploration is to investigate the benefit of LLM API calls for KBQA while limiting the dependence on large volumes of labeled training instances, be it in-domain or few-shot transfer.
At one extreme, we have LLM-free supervised only models.
KB-Binder and gf-LLM demonstrate improvements based on 1 API call per instance.
We have demonstrated that more benefit is obtainable with additional API calls.


While we have verified our individual experimental results by first performing experiments on smaller subsets, the numbers recorded in Tables ~\ref{tab:transfer}, ~\ref{tab:transfer-ablation} and ~\ref{tab:in-domain} are based on single runs for versions of \sys{} as well as for KB-Binder and gf-LLM.
While this understandably affects the reliability of the results, we are hampered by the cost of GPT-4.

Our main experiments are based on datasets built over Freebase as the underlying KB.
Unfortunately, all existing KBQA datasets span only a small set of similar KBs, namely Freebase, Wikidata and DBPedia. 
Creating larger KBQA datasets spanning different types of KBs and domains is a substantial endeavour and we leave it for future work.

\section*{Risks}\label{sec:risks} 

Our work does not carry any obvious risks.

\section*{Acknowledgements}
Mausam is supported by a contract with TCS, grants from IBM, Verisk, Huawei,  Wipro, and the Jai Gupta chair fellowship by IIT Delhi. We thank the IIT-D HPC facility for its computational resources. We also thank Microsoft
Accelerate Foundation Models Research (AFMR) program that helps provide access to OpenAI models.
We are also thankful to Lovekesh Vig for helpful discussions and providing pointers to related work.

\bibliography{main}
\bibliographystyle{acl_natbib}

\appendix
\section{Appendix}

\subsection{\sys{} Implementation Details}
\subsubsection{TIARA and Pangu Retrievers}\label{subsubsection:retriever}
TIARA~\cite{shu-etal-2022-tiara} uses a multi-grain retriever, consisting of an entity retriever, a logical form retriever and a schema retriever.

The {\em entity retriever}, consists of a sequence of three steps - mention detection, candidate generation and entity disambiguation.
Mention detection is performed using span classification, where candidate spans are represented using tokens via BERT, and classified as a mention or not.
The candidate entities for each span are generated using FACC1 alias mapping. 
Entity disambiguation jointly encodes the question and an entity context (entity label and linked relations) in the KB using a cross encoder to get a matching score. 
The entity retriever components require large volumes of data to train.
For every setting, we use a pre-trained entity retriever the GrailQA training set, and fine-tune it on the source using the target few-shots as the validation set.
We use the same entity retriever for Pangu and BERT-Ranking.

The {\em exemplary logical form retriever} enumerates 2-hop paths in the KB from each candidate entity for the question and represents these using s-expression.
Each s-expression are scored by concatenating with the question and encoding using BERT followed by a linear layer.
The parameters are trained using a contrastive objective.

The {\em schema retriever} is independent of the entity retriever.
It scores each entity type (class) and relation in the KB with respect to the question and returns the top entity types and top relations separately.
Each entity type (or relation) is encoded along with the question using a cross-encoder similar to the s-expressions in the logical form retriever.
Entity type and relation classifiers are trained separately.

If a retriever returns data paths in a different representation (e.g. in s-expression as for TIARA), we post-process these and translate to SPARQL using publicly available code.

\subsubsection{Retrieval Re-ranking Prompts}
\label{subsec:prompts_for_retrieval_stage}
We use three different prompts for re-ranking the data paths, entity types and relations in the retrieval result from a supervised retriever.

\textbf{Re-ranking Data Paths:} The prompt for re-ranking data paths uses the following template.


\begin{tcolorbox}[colback=green!5!white, colframe=red!75!black]
   For a given question please select five relevant candidate paths which might be useful for answering it in ranked order. If the number of candidates are five or less then just rerank the candidates. candidate paths are seperated by | symbol. The output should contain only relevant candidates seperated by ``@@''.

\textbf{question: } what the language spoken in indonesia?

\textbf{entity: }  indonesia m.097kp

\textbf{candidate paths: }

(AND language.language\_dialect (JOIN (R language.human\_language.dialects) m.097kp))|

(AND language.language\_dialect (JOIN language.language\_dialect.language m.097kp))| 

\ldots
\end{tcolorbox}

\textbf{Re-ranking Entity Types:} The prompt for re-ranking entity types uses the following template.


\begin{tcolorbox}[colback=green!5!white, colframe=red!75!black]
For a given question please select ten relevant candidate answer entity types which might be useful for answering it in ranked order. If the number of candidates are ten or less then just rerank the candidates. candidate answer entity types are seperated by | symbol. The output should contain only relevant candidates seperated by ``@@''.

\textbf{question:} what the language spoken in indonesia?

\textbf{candidate entity types: }

language.human\_language|

fictional\_universe.fictional\_language | 

\ldots
\end{tcolorbox}

\textbf{Re-ranking Relations:} The prompt for re-ranking relations uses the following template.


\begin{tcolorbox}[colback=green!5!white, colframe=red!75!black]
For a given question please select ten relevant candidate relations which might be useful for answering it in ranked order. If the number of candidates are ten or less then just rerank the candidates. candidate relations are seperated by | symbol. The output should contain only relevant candidates seperated by ``@@''.

\textbf{question:}  what the language spoken in indonesia?

\textbf{candidate Relations:}  

lang.human\_language.countries\_spoken\_in|

location.country.languages\_spoken|

\ldots
\end{tcolorbox}

\subsubsection{Logical Form Generation Prompt}
\label{subsec:prompts_for_generation_stage}

The prompt for generating the logical form uses the following template.


\begin{tcolorbox}[colback=green!5!white, colframe=red!75!black]
Translate the following question to SPARQL for Freebase based on the candidate paths (represented using SPARQL), candidate entities, candidate relations and candidate entity types, which are separated by '|',  retrieved from Freebase by one or more retrievers. Please do not include any other relations, entities and entity types.

Your final SPARQL can have three scenarios: 

1. When you need to just pick from candidate paths represented using SPARQL.

2. When you need to extend one of candidate paths represented using SPARQL using the candidate relations and entity types.

3. When you need to generate a new SPARQL only using the candidate entities, relations and entity types.

For  entity type check please use this relation ``type.object.type''. 

Make sure that the original question can be regenerated only using the identified entity types, specific entities and relations used in the generated SPARQL.

\textbf{\# FEW-SHOTS}

\textbf{question:} when were the texas rangers started

\textbf{candidate entities: } texas rangers m.07l8x

\textbf{candidate paths from Retriever represented using SPARQL: }

SELECT DISTINCT ?x
WHERE \{
ns:m.07l8x ns:sports.sports\_team.founded ?x .
?x ns:type.object.type ns:type.datetime .
\} | \ldots

\textbf{candidate paths from Retriever represented using SPARQL: } 
\ldots

\textbf{candidate relations from Retriever: } 

sports.sports\_team.founded|
military.military\_unit.formed|

\ldots

\textbf{candidate entity types from Retriever: } 
\ldots
media\_common.finished\_work|

transportation.road\_starting\_point|

\ldots

\textbf{SPARQL: }

SELECT DISTINCT ?x

WHERE \{

ns:m.07l8x ns:sports.sports\_team.founded ?x .
\}

\ldots


\end{tcolorbox}


\begin{tcolorbox}[colback=green!5!white, colframe=red!75!black]
\textbf{question:}  what the language spoken in indonesia?

\textbf{candidate entities:} indonesia m.097kp

\textbf{candidate paths from Retriever represented using SPARQL: }

SELECT DISTINCT ?x
WHERE \{

?x ns:lang.language\_family.languages ns:m.097kp .
?x ns:type.object.type ns:language.language\_family .

\} | \ldots

\textbf{candidate relations from Retriever: }

lang.human\_language.countries\_spoken\_in|

location.country.languages\_spoken| \ldots

\textbf{candidate entity types from Retriever: }

language.human\_language|

language.language\_dialect|

\textbf{SPARQL:} 
\end{tcolorbox}

When using a single retriever, we use 5 shots with top-5 retrieval results for data paths and top-10 for relations and entity types. 
When multiple retrievers are used, replicating this for each retriever does not fit in the 8192 context length limit of \texttt{gpt-4-0613}. 
We experimented with two different strategies of reducing the context - reducing the number of exemplars (from 5 to 3 for 2 retrievers) and reducing the retrieval length (top-3 paths and top-5 relations and types for 2 retrievers).
The second strategy worked marginally better over a small evaluation and we used this for our detailed experiments.

\subsubsection{Execution Guided Feedback Prompt}
\label{subsec:prompts_for_execution_based_feedback}

For the execution guided prompt, we use the following template.
In each iteration, we add the LLM output from the previous iterations including the generated logical form to the prompt.

\begin{tcolorbox}[colback=green!5!white, colframe=red!75!black]
The generated SPARQL gives an empty answer when executed on freebase KG, Please generate again a different executable SPARQL using the same context and constraints.

\textbf{SPARQL:}

\ldots
\end{tcolorbox}

\subsection{Supervised Model Details}\label{subsec:supervised_model_details}
All supervised baselines has dataset specific code bases (specifically for data path enumerations,for WebQSP and GrailQA) to handle question types which are not common. For e.g., GrailQA and GraphQA has questions which require comparative and count operators but no such questions are there in the WebQSP dataset. Similarly WebQSP has questions with time-constraints but missing from GrailQA. So GrailQA dataset has significantly broader coverage and more unique canonical logical forms than WebQSP and GraphQA ~\cite{gu:www2021grailqa}. 
So for all supervised baselines, irrespective of the source$\rightarrow$target pairs, we use the respective GrailQA code for fine-tuning in few-shot setting and inference in both zero and few-shot settings with models trained on source. We combine schema elements corresponding to the source and target pairs and use that during zero and few-shot setting.

\paragraph{TIARA:} It consist of four modules - Entity Retrieval, Schema Retriever, Exemplary Logical Form Retrieval and Generator. We use available code and pre-trained models trained on GrailQA and WebQSP \footnote{\url{https://github.com/microsoft/KC/tree/main/papers/TIARA}} (MIT License). 
To adapt these pre-trained models for few-shot transfer, we fine-tune them for one epoch on the target dataset. 
We use learning\_rate of 5e-5 and batch size of 64 for fine-tuning the schema retriever. We use learning\_rate of 3e-5 and batch size of 2 for fine-tuning the exemplary logical form retriever and generator. For all transfer settings, we use corresponding source trained schema retriever to retrieve top-10 relations in zero-shot and fine-tune it on target 100 samples to retrieve top-10 relations in few-shot setting. To get the classes, we use pre-trained model on the source GrailQA  to retrieve top-10 classes for each question in zero-shot setting for GrailQA$\rightarrow$GraphQA and GrailQA$\rightarrow$WebQSP pairs  whereas in few-shot setting  we fine-tune pre-trained model with 100 target samples to get classes for GrailQA$\rightarrow$GraphQA pair. For GrailQA$\rightarrow$WebQSP pair, we use classes predicted by pre-trained model for few-shot setting also. We use relations retrieved by the TIARA schema retriever to get the top-10 classes in zero and few-shot setting for WebQSP$\rightarrow$GrailQA-Tech, WebQSP$\rightarrow$GraphQA-Pop pairs. During path enumerations for GrailQA$\rightarrow$GraphQA and GrailQA$\rightarrow$WebQSP pairs we consider all relations whereas for WebQSP$\rightarrow$GrailQA-Tech, WebQSP$\rightarrow$GraphQA-Pop pairs we ignore relations which starts with ``user.'',``kg.'',``common.'' and ``type.''.

\paragraph{Pangu:} 
We use the existing codebases for Pangu\footnote{\url{https://github.com/dki-lab/Pangu}}.
We use same entity linker as TIARA and fine-tune T5(base) ~\cite{2020t5} based pre-trained discriminator trained on GrailQA and WebQSP datasets for 1 epoch with learning rate of 1e-4 and batch size of 1. We use pre-trained model\footnote{\url{https://github.com/dki-lab/ArcaneQA/tree/main/answer_typing}} on the source GrailQA  to get the top-5 answer types for each question in zero-shot setting for GrailQA$\rightarrow$GraphQA and GrailQA$\rightarrow$WebQSP pairs  whereas in few-shot setting  we fine-tune pre-trained model with 100 target samples to get answer-types for GrailQA$\rightarrow$GraphQA pair. For GrailQA$\rightarrow$WebQSP pair, we use answer-types predicted by pre-trained model for few-shot setting also. We use relations retrieved by the TIARA schema retriever to get the top-5 answer types in zero and few-shot setting for WebQSP$\rightarrow$GrailQA-Tech, WebQSP$\rightarrow$GraphQA-Pop pairs.  

\paragraph{BERT-Ranking} We use existing codebase for BERT-Ranking ~\cite{gu:www2021grailqa}\footnote{\url{https://github.com/dki-lab/GrailQA/tree/main}} (Apache License). 
We fine-tune pre-trained model trained on GrailQA and WebQSP for 1 epoch with learning rate of 1e-3 and batch size of 1. During path enumerations for GrailQA$\rightarrow$GraphQA and GrailQA$\rightarrow$WebQSP pairs we consider all relations whereas for WebQSP$\rightarrow$GrailQA-Tech, WebQSP$\rightarrow$GraphQA-Pop pairs we ignore relations which starts with ``user.'',``kg.'',``common.'' and ``type.''.

For all three models, in the zero-shot setting, we use these for the target without any adaptation.
For few-shot adaptation of Pangu, we fine-tune the discriminator, which is the only trainable component.
TIARA has more trainable components - the logical form ranker and the schema retriever in the retrieval stage and then the generator. 
Each of these can potentially be fine-tuned or left un-adapted. 
Our experiments show that the schema retriever improves on fine-tuning on the target.
Since the generator is trained in pipeline, we fine-tune the generator and the logical-form retriever as well based on this observation.
Similar to our model, we fine-tune a baseline component for the target using 1 training epoch, and for all baselines use the same source-trained entity linker without any adaptation. 

\subsection{KB-Binder and gf-LLM Details}\label{subsec:llm_baseline_details}
For KB-Binder, we make use of publicly available code\footnote{\url{https://github.com/ltl3A87/KB-BINDER}} (MIT License).
We use self-consistency and majority voting with 6 examples, as in the experiments in the paper. 
In the retrieval(-R) setting, KB-Binder samples demonstration examples by retrieving from the entire available training data.
We restrict its retrieval to our target training set $D_t$ with 100 examples. 

For gf-LLM~\cite{shu:arxiv2023bottlenecks}, we remove the notion of synthetic data generation proposed in the paper, so that gf-LLM retrieves KB-context for the question using the same TIARA retriever as our model and provides this as KB-grounding to a LLM for logical form generation in addition to in-context examples. 
Since no code is available for this model, we use our own implementation based on the description in the paper.

KB-Binder and gf-LLM report experiments using \texttt{code-davinci-002} and \texttt{gpt-3.5-turbo-0613} respectively as the LLM.
For consistency and fair comparison, we replace these with \texttt{gpt-4-0613} as in our approach.
These models generate logical forms in s-expression, which we preserve.
As in our approach, for the zero-shot setting we provide exemplars to both models from the source.

For fair comparison, we use \texttt{gpt-4-0613} as LLM for KB-Binder, gf-LLM and FuSIC-KBQA. For FuSIC-KBQA we use $temperature=0$ and $n=1$.

\subsection{Dataset Construction Details}\label{subsec:dataset_construction_details}

For GrailQA-Tech, we include questions from the following domains from the GrailQA train and dev set: computer, medicine, astronomy, biology, spaceflight.

For GraphQA-Pop, we include questions from the GraphQA test and train sets: cvg, food, business, aviation, comic books, religion, royalty, architecture, exhibitions, distilled-spirits. 

\subsection{Compute Infrastructure and Time Taken}\label{subsec:compute_details}
We use Hugging Face ~\cite{wolf-etal-2020-transformers}, PyTorch ~\cite{pytorch}  for our experiments and use the  Freebase setup specified on github \footnote{\url{https://github.com/dki-lab/Freebase-Setup}}. We use NVIDIA A100 GPU with 40 GB GPU memory and 32 GB RAM for fine-tuning of different models. Fine-tuning of different models takes less than an hour. 


On average, for each test instance, retrieval (with one retriever) takes 14.4 secs, retrieval re-ranking 9.6 secs, logical form generation 14.4 secs and EGF 18.9 secs.

\subsection{Few-shot Transfer Experiments: Retrieval Re-ranking analysis}\label{subsec:few_shot_transfer_details}

Beyond logical form evaluation, we also directly evaluated retrieval performance for the different versions of \sys{}.
These clearly demonstrate that while \sys{}(P) has much poorer retrieval performance than \sys{}(T), \sys{}(T,P) is better than both, validating the use of multiple retrievers.
Further, LLM re-ranking always significantly boosts retrieval performance.

We use the recall error rate to measure the retrieval performance which is defined as the \% of questions having recall $<$1.0 in terms missing gt KB elements (excluding mentioned entities) from retrieval output i.e.,lower is better). As in Tab.~\ref{tab:retrieval-reranking}, \sys{(T)} has low recall error rate as compared to the TIARA across all source and target pairs and in both zero and few-shot setting. \sys{(P)} has same recall error rate as Pangu (except for WebQSP$\rightarrow$GraphQA-Pop) because number of retrieved relations and types are mostly $\leq$10. And same is true for data paths$\leq$5. \sys{(T,P)} with  TIARA and Pangu retrievers has lower recall error rate as compared to \sys{(T)} and \sys{(P)} in zero-shot and few-shot setting. And \sys{(T)} outperforms \sys{(P)} in all cases. This also correlates with the KBQA performance mentioned in the Tab. \ref{tab:transfer} (except for GrailQA$\rightarrow$GraphQA where \sys{(P)} is better than \sys{(T)} for both zero-shot and few-shot.) 

\subsection{Transfer Experiment with Microsoft Academic KG}\label{subsec:makg_details}
Due to smaller number of schema elements in Microsoft Academic KG, a supervised retriever that returns the top-K schema elements is no longer necessary. We instead directly rerank all schema elements to obtain top-k. 
For gf-LLM, we follow the approach as specified in the paper, selecting top-k schema elements directly from the supervised retriever. 
We use gpt-4-0613 as LLM for both FuSIC-KBQA and gf-LLM.
We do not perform constrained decoding during generation for TIARA.

\subsection{In-domain Experiments}\label{subsec:in_domain_details}
\paragraph{TIARA:} It consists of four modules - Entity Re-trieval, Schema Retriever, Exemplary Logical Form Retrieval and Generator. We use pre-trained models\footnote{\url{https://kcpapers.blob.core.windows.net/tiara-emnlp2022/TIARA_DATA.zip}} corresponding to 100\% training data and use existing codebase\footnote{\url{https://github.com/microsoft/KC/tree/main/papers/TIARA}} to train on 5\% and 10\% training data. We train a mention detection model for 50 epochs and choose the best model based on the validation set. We use a non-bootstrapped strategy for sampling negative logical forms during the training of Exemplary Logical Form Retrieval. 
\paragraph{Pangu:} 
We use the existing codebase for Pangu\footnote{\url{https://github.com/dki-lab/Pangu}} to train a model on 5\% and 10\% training data. And use pre-trained model\footnote{\url{https://github.com/dki-lab/Pangu/blob/main/trained_models.md}} corresponding to 100\% training data. We use the same entity linker as TIARA and use T5(base) ~\cite{2020t5}  as our base model.
\paragraph{RetinaQA:} 
We use the existing codebase for RetinaQA\footnote{\url{https://github.com/dair-iitd/RetinaQA}} to train a sketch generator and discriminator on 5\% and 10\% training data. And TIARA codebase to train  schema and logical form retriever on 5\% and 10\% training data. We use pre-trained models for  sketch generator and discriminator corresponding to the 100\% training data. We use the same entity linker as TIARA.

\subsection{Equivalence Check for SPARQL}\label{subsec:em-sparql}
We extract classes, relations (including reverse), entities, literals and functions from predicted and gold SPARQL. We use lookup in ontology files to extract classes, relations and regex to extract literals and entities. And for a given question we perform two types of matching between elements extracted from predicted and gold SPARQL  (i) Multi-set based exact match: two multi-sets should be same and (ii) Set based exact match: set of elements excluding literals should be same. So if (i) condition is true and answer F1 is 1.0 then EM is 1.0 whereas if (ii) is false then EM is 0.0 and in all other cases predicted SPARQL is reviewed manually.
\subsection{Experiments with Mixtral} \label{subsec:open-source-llm}
We use \texttt{Mixtral 8x7B Instruct v0.1} as open-source llm for FuSIC-KBQA with $temperature=1e-5$. We use same prompt as LLM for retrieval re-ranking and execution guided feedback. We also use same prompt and few-shots for logical form generation except we enclose few-shots within ``\#\#\#\# Here are some examples:   \#\#\#\#'' delimiters and indicate the start of test query via ``$<<<>>>$''.

\begin{tcolorbox}[colback=green!5!white, colframe=red!75!black]
Translate the following question \textbf{after $<<<>>>$} to SPARQL for Freebase based on the candidate paths (represented using SPARQL), candidate entities, candidate relations and candidate entity types, which are separated by '|',  retrieved from Freebase by one or more retrievers. Please do not include any other relations, entities and entity types.

Your final SPARQL can have three scenarios: 

1. When you need to just pick from candidate paths represented using SPARQL.

2. When you need to extend one of candidate paths represented using SPARQL using the candidate relations and entity types.

3. When you need to generate a new SPARQL only using the candidate entities, relations and entity types.

For  entity type check please use this relation ``type.object.type''. 

Make sure that the original question can be regenerated only using the identified entity types, specific entities and relations used in the generated SPARQL.\\
\textbf{\#\#\#\#} \\
\textbf{\# FEW-SHOTS} \\
\textbf{``Here are some examples:''} \\
\textbf{question:} when were the texas rangers started

\textbf{candidate entities: } texas rangers m.07l8x

\textbf{candidate paths from Retriever represented using SPARQL: }

SELECT DISTINCT ?x
WHERE \{
ns:m.07l8x ns:sports.sports\_team.founded ?x .
?x ns:type.object.type ns:type.datetime .
\} | \ldots

\textbf{candidate paths from Retriever represented using SPARQL: } 
\ldots

\textbf{candidate relations from Retriever: } 

sports.sports\_team.founded|
military.military\_unit.formed|

\ldots

\textbf{candidate entity types from Retriever: } 
\ldots
media\_common.finished\_work|

transportation.road\_starting\_point|

\ldots


\end{tcolorbox}


\begin{tcolorbox}[colback=green!5!white, colframe=red!75!black]

\textbf{SPARQL: }

SELECT DISTINCT ?x

WHERE \{

ns:m.07l8x ns:sports.sports\_team.founded ?x .
\}

\ldots \\
\textbf{\#\#\#\#} \\
\textbf{$<<<>>>$} \\
\textbf{question:}  what the language spoken in indonesia?

\textbf{candidate entities:} indonesia m.097kp

\textbf{candidate paths from Retriever represented using SPARQL: }

SELECT DISTINCT ?x
WHERE \{

?x ns:lang.language\_family.languages ns:m.097kp .
?x ns:type.object.type ns:language.language\_family .

\} | \ldots

\textbf{candidate relations from Retriever: }

lang.human\_language.countries\_spoken\_in|

location.country.languages\_spoken| \ldots

\textbf{candidate entity types from Retriever: }

language.human\_language|

language.language\_dialect|

\textbf{SPARQL:} 
\end{tcolorbox}

\begin{table*}[ht]
\begin{center}
\small
            \begin{tabular}[b]{lllcccc}
\hline
& & \multirow{2}{4em}{\textbf{Model}}   & \textbf{GrailQA$\rightarrow$} &\textbf{GrailQA$\rightarrow$} &\textbf{WebQSP$\rightarrow$} &\textbf{WebQSP$\rightarrow$}\\  
& &  & \textbf{GraphQA} &\textbf{WebQSP} &\textbf{GrailQA-Tech} &\textbf{GraphQA-Pop}\\  
\hline
\multirow{5}{1em}{\rotatebox[origin=c]{90}{Zero-shot}} & 
\multirow{2}{5em}{Supervised}  & TIARA &17.2 & 35.6&8.6  &22.0\\
& & Pangu & 37.0 & 48.0 &47.8&67.8 \\
& \multirow{3}{5em}{Fusion}  & \sys{}(T) &12.4 & 25.2&3.2  &13.4\\ 
  &  & \sys{}(P) & 37.0& 48.0& 47.8 &67.4\\
  &  & \sys{}(T,P) &\textbf{9.6} &\textbf{23.6} & \textbf{2.6} &\textbf{10.2}\\
\hline
 \multirow{5}{1em}{\rotatebox[origin=c]{90}{Few-shot}} & 
 \multirow{2}{5em}{Supervised}  & TIARA &17.6 & 30.8&5.0  &24.2\\
& & Pangu & 31.8 & 43.8 &27.0&56.0 \\

 & \multirow{3}{5em}{Fusion}   & \sys{}(T) &10.8 & 23.8&3.2  &13.4\\ 
  &  & \sys{}(P) &31.8 &43.8 & 27.0&54.8\\
  &  & \sys{}(T,P) & \textbf{7.6}& \textbf{21.6}& \textbf{2.8} &\textbf{9.2}\\

\hline
\end{tabular}
\end{center}
\caption{Retrieval performance using error rate (lower is better) in few-shot and zero-shot transfer learning settings  
of different models for different source$\rightarrow$target pairs, evaluated over 500 randomly sampled instances over the test sets. \sys{}(T), \sys{}(P), \sys{}(T,P) denote \sys{} used with TIARA, Pangu, and TIARA and Pangu together as supervised retrievers. 
}
\label{tab:retrieval-reranking}
\end{table*}

\end{document}